\documentclass[authoryear, final,times,twocolumn]{elsarticle}

\usepackage{amssymb}
\usepackage{comment}
\usepackage{amsmath}
\usepackage{caption}
\usepackage{hyperref}
\usepackage{natbib}
\journal{Pattern Recognition Letters}

\begin{document}

\begin{frontmatter}



\title{MARS: Magnitude-Aware Rank Statistics}

\author[1]{Muhammad Rajabinasab} 
\affiliation[1]{organization={University of Southern Denmark},
            city={Odense},
            country={Denmark}}
\author[1]{Afsaneh M. Nejad}         
\author[1]{Arthur Zimek}

\begin{abstract}
Comprehensive evaluation of machine learning models is the key to make sure that they perform as robustly and consistently as desired. In order to summarize the experimental results and pick a winner, Critical Difference (CD) diagrams are used. Standard CD diagrams rely on discrete ranks, discarding the magnitude of performance gaps between models, raising an issue which we call magnitude-blindness. In order to address this issue, we propose Magnitude-Aware Rank Statistics (MARS) that incorporates a relative margin coefficient as a weight for the discrete ranks. This coefficient scales ranks based on the distance between the best and worst performers, with a dynamic projection to handle boundary cases. Followed by the calculation of a CD value, MARS results in a more realistic statistical representation of differences of model performances and more insights on how methods actually perform in vast and extensive experimental settings.  \\\\
\emph{Preprint submitted to Elsevier Pattern Recognition Letters.}
\end{abstract}




\end{frontmatter}



\section{Introduction}\label{sec:intro}
The comprehensive and rigorous evaluation of machine learning models serves as the cornerstone of algorithmic progress, providing the empirical basis upon which novel methods and algorithms are validated and developed. Many different metrics are proposed to measure different aspects of performance for different tasks and applications~\citep{Ferri2009}. Novel developments require an extensive evaluation and analysis in order to demonstrate reliable performance. This is a necessity to ensure an algorithm's performance is robust, but on the other hand, it makes the comparison task and picking \emph{the final winner} very difficult. Here is where rank statistics methods play a major role. Traditional approaches, most notably the Friedman test followed by Nemenyi or Bonferroni-Dunn post-hoc analyses \cite{demsar2006}, have long been the gold standard for comparing multiple classifiers over multiple datasets.
What these methods do is basically averaging the ranking of each method against the competitors based on different observations and calculating the Critical Difference (CD) value, to define what amount of difference in the average ranking is statistically significant. However, these methods rely on standard rank aggregation, which reduces complex performance distributions to ordinal integers. This methodology introduces a significant oversimplification that we could call \emph{magnitude-blindness}, treating a marginal victory of $0.01\%$ as equivalent to a clear dominance of $50\%$.

In high-stakes domains, this loss of signal is not merely a statistical nuance; it can mask catastrophic failure modes that carry significant real-world consequences. 
In medical diagnostics, for instance, an algorithm’s frequent marginal superiority in common cases may obscure life-threatening failures in rare but critical pathologies \citep{Kelly2019, Rajpurkar2022, DeGrave2021}. Such rank-based assessments often ignore the \emph{calibration} of the model, where a system may be correct in its prediction, but not with a high confidence, or dangerously overconfident when incorrect \citep{Guo2017}. Similarly, in the development of autonomous systems, standard rank-based metrics often fail to distinguish between minor navigational errors and high-magnitude safety violations that lead to system collisions \citep{Koopman2017, Codevilla2018, Philion2020}. This is particularly perilous in long-tail scenarios, where the statistical rarity of a failure does not diminish its physical severity \citep{Kalra2016}. Even in non-high-stake domains, an evaluation which is summed up with an approach suffering from \emph{magnitude-blindness} might lead to unrealistic conclusions. For example, in Large Language Model (LLM)~\citep{Zhao2023LLMSurvey} benchmarking, a model might achieve a higher winning rate by providing stylistically pleasing responses to trivial queries while failing catastrophically on complex reasoning tasks, a phenomenon often described as \emph{reward hacking} \citep{Amodei2016, Skalse2022}. By ignoring the distance between the winner and the loser, we risk incentivizing volatility to achieve frequent marginal superiority over fewer, but major improvements.

However, performance gaps are not necessarily comparable over different datasets. This is the main motivation for a rank-based analysis.
Consequently, there is a critical need for a framework that preserves the stability and the compactness of rank-based comparisons while remaining sensitive to the \emph{cardinal utility} of (normalized) performance gaps \citep{Sen1970, Dwork2001}. Such a framework must ensure that \emph{robustness} is not sacrificed for an overly simplistic concept such as \emph{winning frequency}, effectively bridging the gap between ordinal stability and metric sensitivity.

The reliance on traditional post-hoc procedures such as the Nemenyi \citep{demsar2006} tests, imposes a significance threshold that is fundamentally indifferent to the distribution and the magnitude of performance gaps. In these methods, either a single Critical Difference (CD) value or a rank-sum $p$-value determines the boundaries of statistical equivalence. Wilcoxon-Holm~\citep{holm1979simple} incorporates metric values to calculate $p$-values and CD, however the ranks are still \emph{magnitude-blind}. This creates a paradox in comparative analysis: a model that consistently achieves marginal victories is statistically rewarded the same as a model that demonstrates a very high performance superiority. Furthermore, because these tests operate in a discrete ordinal space, the resulting \emph{statistical significance} often fails to expand in response to the actual stability of the performance margins. This rigidity can lead to a fake sense of stability, where two models are declared statistically equal, with no significant statistical difference, simply because their win-loss frequency is balanced, even if one model’s losses are catastrophic while its wins are negligible.

In this paper, we propose Magnitude-Aware Rank Statistics (MARS)\footnote{Python code available: {\url{https://github.com/mrajabinasab/MARS/}}}. Unlike the traditional approaches by \citet{demsar2006}, MARS integrates the performance metric values into the ranking process, transforming discrete ordinal ranks into continuous magnitude-aware rank scores. This transformation ensures that the magnitude of superiority is preserved. Furthermore, we introduce a dynamic regularization of the Critical Difference, which scales the significance threshold based on the observed volatility of the performance gaps. MARS employs a non-parametric permutation test to evaluate the stability of these scores, moving beyond the rigid assumptions of standard post-hoc tests. By reconciling the stability of rank-based visualization with the sensitivity of metric-based analysis, MARS offers a more transparent and reliable framework for identifying the \emph{true winner} in large-scale performance comparisons.

In the remainder, we review the traditional rank statistics and critical difference (Section~\ref{sec:background}). We present MARS, its methodology, its formal definition, and the corresponding calculations (Section~\ref{sec:prop}). We discuss comparative scenarios in which the traditional approach fails, but MARS succeeds in providing \emph{realistic}, \emph{correct}, and additional insights (Section~\ref{sec:why}). We conclude the paper (Section~\ref{sec:conc}), providing a short summary and mentioning limitations which can lead to future research.

\section{Background}\label{sec:background}
The comprehensive evaluation of machine learning models requires robust statistical tests to provide a valid comparison across heterogeneous datasets \citep{Japkowicz2025}. This process typically follows a two-stage frequentist pipeline: a global omnibus test to detect any significant differences within the group, followed by post-hoc pairwise analyses to identify specific winning relationships, often visualized through Critical Difference (CD) diagrams.

The procedure begins by converting the raw performance metrics values (e.g., accuracy) into ordinal ranks for each dataset, where the best-performing model is assigned rank $1$ and the worst rank $k$. These ranks are averaged across all $N$ datasets to produce a single ranking value which is globally meaningful. Once these ranks are established, the Friedman test \citep{friedman1937use} is used to determine if any statistically significant differences exist among the group. If the null hypothesis is rejected, the analysis proceeds to a post-hoc evaluation to identify specific winning relationships. This is traditionally achieved via the Nemenyi test \citep{demsar2006}, which calculates a global Critical Difference (CD), a fixed distance that any two models must exceed to be considered significantly different. Alternatively, post-hoc pairwise comparisons using the Wilcoxon signed-rank test can be used. This test evaluates every possible pair of classifiers to determine if one consistently outperforms the other based on the magnitude and direction of their performance differences. To prevent the accumulation of Type I errors, where random noise is mistaken for a significant discovery, the resulting $p$-values are filtered through Holm’s step-down procedure. This method sequentially compares each $p$-value against a progressively stricter significance threshold, ensuring that only the most robust differences are retained. 

In this section, we dive deeper into each one of these steps to develop a better understanding of their underlying mechanism.

\subsection{The Friedman Test}
After calculating the average ranks and before performing pairwise comparisons, a non-parametric equivalent of the repeated-measures ANOVA, the Friedman test \citep{friedman1937use, friedman1940comparison}, must be conducted. It tests the null hypothesis that all $k$ classifiers perform equally. The Friedman statistic is calculated based on the average ranks $R_j$:
\begin{equation}
\chi_F^2 = \frac{12N}{k(k+1)} \left[ \sum_{j=1}^k R_j^2 - \frac{k(k+1)^2}{4} \right]
\end{equation}
If the resulting $p$-value is below a pre-specified significance level $\alpha$ (typically $0.05$), the null hypothesis is rejected, justifying the use of post-hoc tests.

\subsection{The Nemenyi Test}
As discussed in the foundational work of \citet{demsar2006}, the Nemenyi test is a post-hoc procedure that calculates the average rank $R_j = \frac{1}{N} \sum_i r_i^j$ for each classifier. Two classifiers are considered significantly different if their average ranks differ by at least the Critical Difference (CD), defined as:
\begin{equation}
\mathit{CD} = q_{\alpha} \sqrt{\frac{k(k+1)}{6N}}
\end{equation}
where $q_{\alpha}$ is the Studentized range statistic for a significance level $\alpha$. While computationally elegant and widely adopted for its visual simplicity in CD diagrams, this approach is known to be conservative: Performing all possible ${m = k(k-1)/2}$ pairwise comparisons simultaneously using a single global threshold, it often fails to detect significant differences in scenarios with a small number of datasets \citep{garcia2008extension}.

\subsection{Wilcoxon Signed-Rank and Holm’s Procedure}
A more powerful alternative involves pairwise Wilcoxon signed-rank tests. Unlike the Nemenyi test, this procedure evaluates each pair of classifiers independently by ranking the absolute differences in their performance metrics across $N$ datasets. To control the Family-Wise Error Rate (FWER) across multiple comparisons, $p$-values are adjusted using Holm’s step-down procedure \citep{holm1979simple}, following a sequential logic:
\begin{enumerate}
    \item All $m$ pairwise $p$-values are sorted in ascending order: $p_1 \le p_2 \le \dots \le p_m$.
    \item Each $p_i$ is compared to a dynamic adjusted significance level $\alpha_i = \frac{\alpha}{m - i + 1}$.
    \item The null hypothesis for a pair is rejected only if $p_i \le \alpha_i$. The process continues until the first $i$ where $p_i > \alpha_i$, at which point all remaining hypotheses are retained.
\end{enumerate}

There is no global CD in Wilcoxon-Holm. If the specific pairwise test fails to find a winner after the Holm adjustment, the two methods will be deemed as statistically similar.

While more sensitive than the Nemenyi test, this pipeline still operates fundamentally on the discrete rank space $\{1, \dots, k\}$. By converting performance into ordinal ranks, it inherently ignores the magnitude of the underlying metric values, such as Accuracy or AUC \citep{benavoli2016should}, potentially masking the distinction between a marginal victory and a substantial performance gain.
\section{Magnitude-Aware Rank Statistics}\label{sec:prop}
In order to address \emph{magnitude blindness}, we propose Magnitude-Aware Rank Statistics (MARS), a weighted ranking scheme where the influence of a rank is scaled by the relative difference of the observation to the best-performing method. Thus, MARS takes into account the performance gaps between methods across the observations.

\subsection{Calculating Magnitude-Aware Rank Scores}
Let $y_{i,j}$ be the value of a performance metric for method $j$ on dataset $i$. Let $y_{i,\max}$ and $y_{i,\min}$ be the maximum and minimum observed values on dataset $i$, respectively. For any method $j$ where ${y_{i,j} > y_{i,\min}}$, the weight coefficient $w_{i,j}$ is defined as:
\begin{equation}
w_{i,j} = \frac{y_{i,\max} - y_{i,\min}}{y_{i,j} - y_{i,\min}}
\end{equation}
To avoid division by zero when $y_{i,j} = y_{i,\min}$, we implement a dynamic penalty based on the observed distribution of the other methods.
Let the sorted weights for methods on dataset $i$ that are strictly greater than the minimum be $W_i = \{w_{i,(1)}, \dots, w_{i,(k-m)}\}$, where $m$ is the number of methods tied at the minimum value. We define the weight for these minimum-value methods as:
\begin{equation}
w_{i,\min} = w_{i,(k-m)} + \Delta w_{\max}
\end{equation}
where:
\begin{equation}
\Delta w_{\max} = \max_{r < k-m} (w_{i,(r+1)} - w_{i,(r)})
\end{equation}
If all methods except the minimum are equal, or $k-m=1$, $\Delta w_{\max}$ defaults to $w_{i,(1)}$. This ensures the worst performer is penalized relative to the most significant performance difference observed in the dataset.

The final rank for the method $j$ is the mean weighted rank across all $N$ datasets:
\begin{equation}
\hat{R}_j = \frac{1}{N} \sum_{i=1}^N (r_{i,j} \cdot w_{i,j})
\end{equation}
where $r_{i,j}$ is the standard discrete rank ($1$ for best, $k$ for worst).

\subsection{MARS Critical Difference}
The MARS rank scores theoretically lie within $[1, \infty)$. To account for the shift in scale compared to standard ranks in the range $[1, k]$, we adjust the critical difference:
\begin{equation}
\mathit{CD}_{\mathit{MARS}} = q_\alpha \sqrt{\frac{k(k+1)}{6N}} \cdot \frac{\sigma(R_\mathit{MARS})}{\sqrt{\frac{k^2 - 1}{12}}}
\end{equation}
where:
\begin{itemize}
    \item $q_\alpha$ is the critical value for the Nemenyi test at significance level $\alpha$;
    \item $k$ is the number of compared methods;
    \item $N$ is the number of datasets;
    \item $\sigma(R_\mathit{MARS})$ is the empirical standard deviation of the MARS-weighted ranks;
    \item $\sqrt{\frac{k^2 - 1}{12}}$ represents the theoretical standard deviation of standard integer ranks, derived from properties of the discrete uniform distribution \citep{walpole2012}.
\end{itemize}

The standard rank statistics assumes a uniform distance between ranks. MARS' weighted approach transforms the rank space into a metric space that reflects performance density. The inclusion of the performance metric value in the calculation of the MARS' score introduces a notion of relativity. The weights which transform the discrete ranks into the continuous metric space, penalize a method based on the closeness of its performance to the worst performance figure in a specific observation (e.g., dataset). After the averaging step, this metric value incorporation yields a statistically meaningful depiction of both how a method is ranked, and how decisive the win has been. Alongside the adjusted CD value, this allows MARS to provide more comprehensive and yet condensed insights into a vast and extensive empirical evaluation. Section~\ref{sec:why}, we will demonstrate how this approach helps MARS to provide more informative analyses in different scenarios. 

\subsection{Significance Testing}
To determine the statistical significance of the observed weighted ranks, we employ a non-parametric permutation test as the global test, alongside the Wilcoxon-Holm procedure~\citep{holm1979simple}. Wilcoxon-Holm process uses metric values to determine the $p$-values and the dynamic pairwise Critical Difference. As we calculate our own Critical Difference value, we only use Wilcoxon-Holm for the $p$-value testing, alongside an additional global permutation test. The null hypothesis $H_0$ assumes that there is no inherent difference between the compared methods, and any observed variance in the average weighted ranks is due to chance.

The test statistic is defined as the variance of the average weighted ranks:
\begin{equation}
\mathcal{S} = \text{Var}(\hat{R}_1, \hat{R}_2, \dots, \hat{R}_k)
\end{equation}
A high variance indicates a strong separation between the methods, suggesting that some methods consistently outperform others in a magnitude-aware sense. To estimate the distribution of $\mathcal{S}$ under $H_0$, we perform $\rho$ permutations (typically $\rho=10000$). In each iteration, the performance values $y_{i,j}$ are randomly shuffled within each dataset $i$, and the weighted rank matrix is recomputed to yield a null statistic $\mathcal{S}^*$.

The $p$-value is then calculated as the proportion of permutations where the null statistic is greater than or equal to the observed statistic:
\begin{equation}
p = \frac{1}{\rho} \sum_{p=1}^\rho \mathbb{I}(\mathcal{S}^*_p \geq \mathcal{S})
\end{equation}
where $\mathbb{I}(\cdot)$ is the indicator function. The resulting permutation $p$-value serves as a global indicator of experimental validity. We reject the null hypothesis of model equivalence only if $p < \alpha$ (typically $\alpha = 0.05$). If this condition is met, the observed variance in magnitude-aware rank scores is considered statistically significant, thereby justifying the subsequent pairwise comparison via the $\mathit{CD}_{\mathit{MARS}}$ threshold.

This approach ensures that the significance level is tied directly to the empirical distribution of the weighted ranks rather than assuming a standard distribution, providing a robust frequentist foundation for the MARS analysis. It is noteworthy that as the metric values are incorporated in the calculation of the \emph{global} significance, it is expected to be more sensitive to the number of observations compared to standard rank statistics. It is therefore advised to apply a strict rejection only using the Wilcoxon-Holm process, and to use the $p$-value permutation testing yields as complementary information.

\section{Experiments and Empirical Analysis}
\label{sec:why}

To demonstrate the efficacy of the MARS framework, we define six synthetic experimental scenarios. These scenarios are specifically designed to highlight common edge cases in machine learning benchmarking where standard integer-based rank analysis fails to provide a truthful representation of model performance. In each case, we compare the traditional Friedman-Wilcoxon-Holm pipeline as the strongest standard rank statistics against the MARS-adjusted Critical Difference (MCD) to observe how magnitude-awareness alters, and improves the statistical conclusions.

\begin{table}[tb!] 
\centering
\caption{Average Accuracy for Scenario 1}
\label{tbl:sc1}
\begin{tabular}{lcc}
\hline
\textbf{Classifier} & \textbf{$D_{0-19}$ (Major Win)} & \textbf{$D_{20-39}$ (Minor Loss)} \\ \hline
Method A            & 0.95                            & 0.94                              \\
Method B            & 0.50                            & 0.95                              \\
Method C            & 0.30                            & 0.30                              \\ \hline
\end{tabular}

\vspace{0.5cm}

\includegraphics[width=.82\columnwidth]{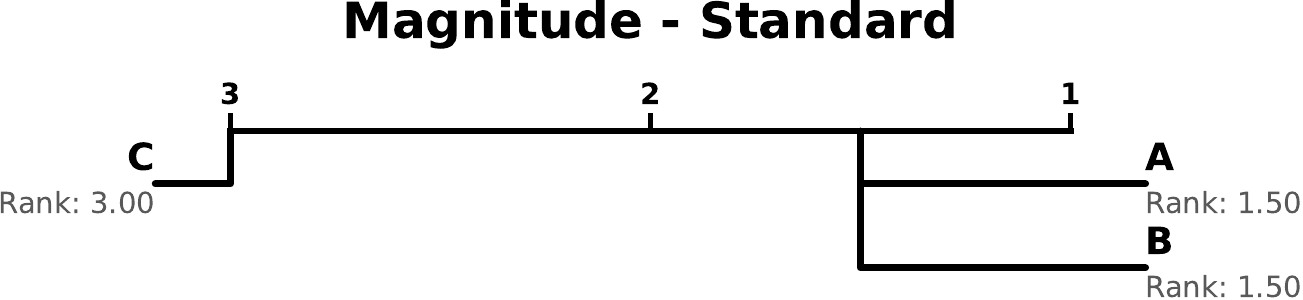}
\vspace{0.15cm}\\
\includegraphics[width=\columnwidth]{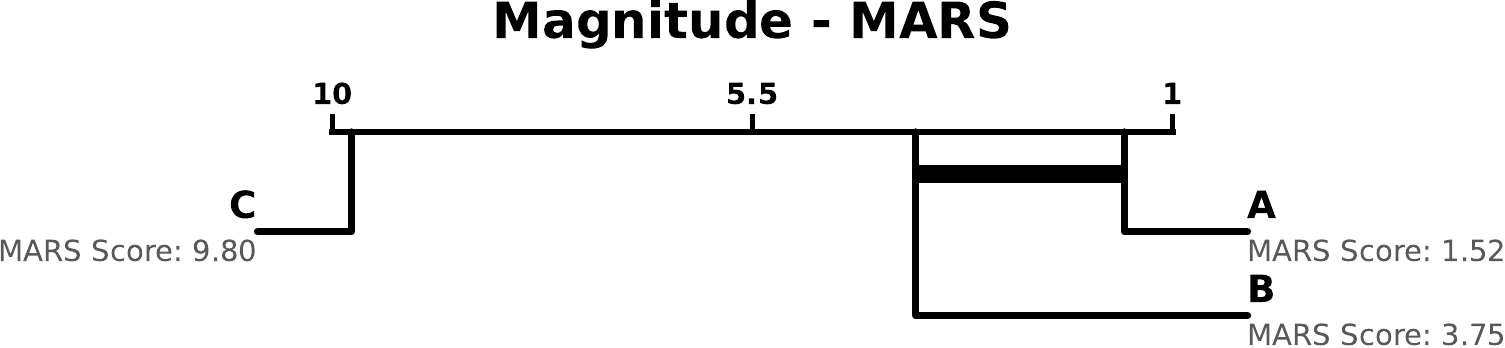}
\captionof{figure}{Comparison for Scenario 1: Standard (Top) vs.~MARS (Bottom).}
\label{fg:sc1}
\end{table}

\subsection{Scenario 1: Magnitude of Performance}
This scenario evaluates the impact of decisive victories versus marginal wins. In the first 20 datasets, Method A outperforms Method B by a large margin. In the subsequent 20 datasets, Method B wins, but only by a negligible margin ($0.01\%$). Method C consistently shows an inferior performance. The details of this scenario are shown in Table~\ref{tbl:sc1}, the resulting CD diagrams are presented in Figure~\ref{fg:sc1}. In the standard CD diagram, the rank value is the average rank of each method over all observations, while in MARS, it is the Magnitude-Aware Rank Statistics Score, as introduced in Section~\ref{sec:prop}.

\textbf{Analysis:} Standard rank analysis treats a win as a win regardless of the gap, resulting in Method A and B appearing statistically indistinguishable. MARS, however, applies a penalty to Method B for its catastrophic performance in the first half, correctly isolating Method A as the superior model.

\begin{table}[tb!]
\centering
\caption{Average Accuracy for Scenario 2}
\label{tbl:sc2}
\begin{tabular}{lcc}
\hline
\textbf{Classifier} & \textbf{$D_{0-29}$ (Consistent)} & \textbf{$D_{30-39}$ (Failure)} \\ \hline
Method A            & 0.81                             & 0.10                           \\
Method B            & 0.80                             & 0.95                           \\
Method C            & 0.20                             & 0.08                           \\ \hline
\end{tabular}

\vspace{0.5cm}

\includegraphics[width=.82\columnwidth]{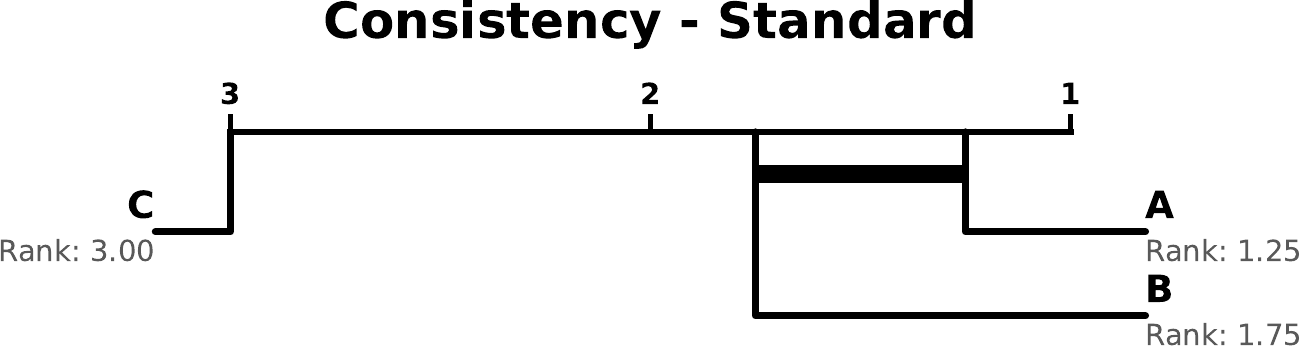} \\
\vspace{0.5cm}
\includegraphics[width=\columnwidth]{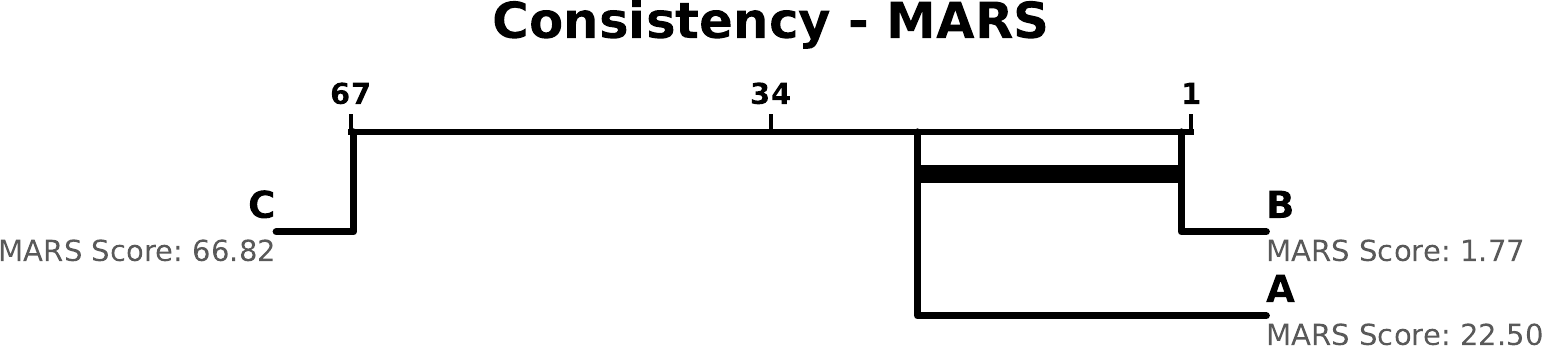}
\captionof{figure}{Comparison for Scenario 2: Standard (Top) vs.~MARS (Bottom).}
\label{fg:sc2}
\end{table}

\subsection{Scenario 2: Consistency and Robustness}
Scenario 2 simulates the \emph{Winner's Curse}. Method A wins in $75\%$ of datasets by a small margin, but fails by a large margin in the remaining $25\%$. Method B remains highly performant and robust across all tasks. The details of this scenario are shown in Table~\ref{tbl:sc2}, the resulting CD diagrams are presented in Figure~\ref{fg:sc2}.

\textbf{Analysis:} While standard ranking rewards Method A's win frequency, the MARS-adjusted Critical Difference (MCD) stretches the rank space to account for the massive variance in A's performance, ultimately favoring the robustness of Method B. It also penalizes the catastrophic performance of A in 25\% of the cases highly, assigning it to the same category as C as an unreliable method.

\begin{table}[tb!]
\centering
\caption{Average Accuracy for Scenario 3}
\label{tbl:sc3}
\begin{tabular}{lcc}
\hline
\textbf{Classifier} & \textbf{$D_{0-19}$ (Volatile Win)} & \textbf{$D_{20-39}$ (Volatile Crash)} \\ \hline
Steady A            & 0.70                                & 0.70                                  \\
Volatile B          & 0.90                                & 0.10                                  \\
Method C            & 0.50                                & 0.50                                  \\ \hline
\end{tabular}

\vspace{0.5cm}

\includegraphics[width=.82\columnwidth]{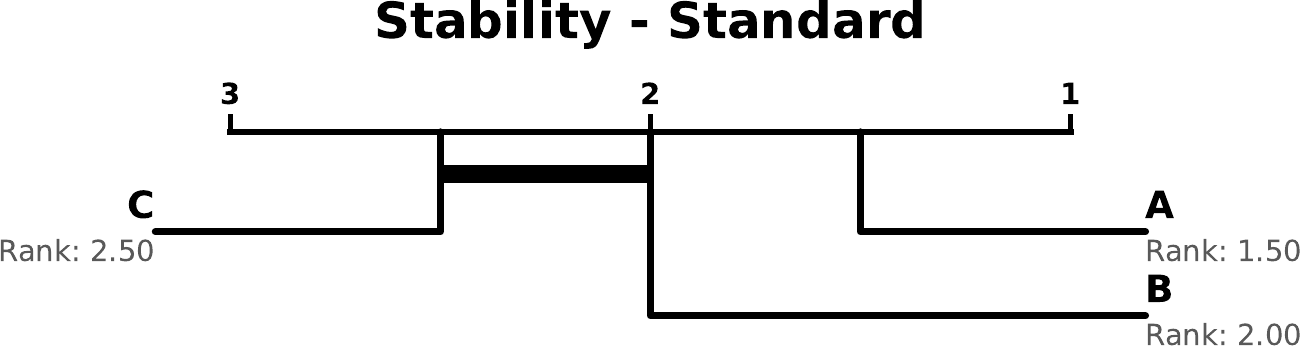} \\
\vspace{0.5cm}
\includegraphics[width=\columnwidth]{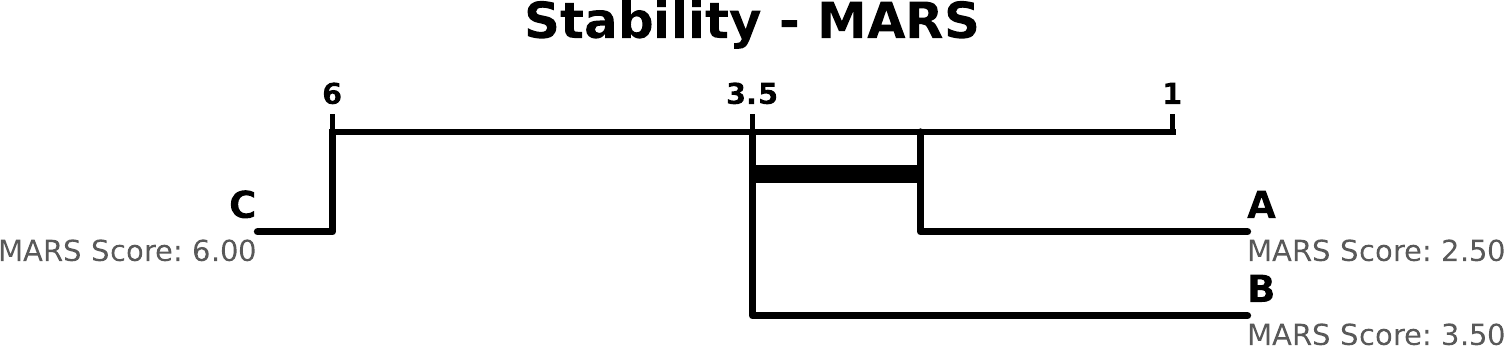}
\captionof{figure}{Comparison for Scenario 3: Standard (Top) vs.~MARS (Bottom).}
\label{fg:sc3}
\end{table}

\subsection{Scenario 3: Stability against Volatility}
We compare a Steady performer (A) against a Volatile performer (B). Method B wins 50\% of the time but crashes to near-zero accuracy in the other half, whereas Method A maintains consistent competitiveness. The details of this scenario are shown in Table~\ref{tbl:sc3}, and the resulting CD diagrams are presented in Figure~\ref{fg:sc3}.

\textbf{Analysis:} Standard analysis often groups these methods in the same significance clique. MARS separates them, penalizing Volatile B for its lack of reliability while keeping Steady A as the top rank. It also separates B and C successfully, as B, despite volatility, shall be deemed as more reliable at least for a subset of the problems.

\begin{table}[tb!]
\centering
\caption{Average Accuracy for Scenario 4}
\label{tbl:sc4}
\begin{tabular}{lcc}
\hline
\textbf{Classifier} & \textbf{$D_{0-29}$ (Noise)} & \textbf{$D_{30-39}$ (Significant)} \\ \hline
Method A            & 0.8499                             & 0.90                                   \\
Method B            & 0.8500                             & 0.15                                   \\
Method C            & 0.7500                             & 0.75                                   \\ \hline
\end{tabular}

\vspace{0.5cm}

\includegraphics[width=.82\columnwidth]{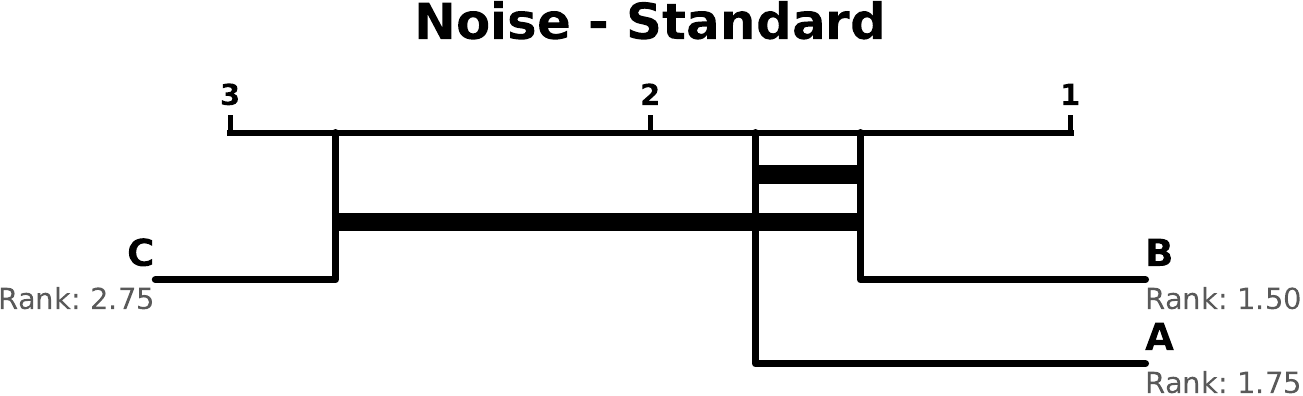} \\
\vspace{0.5cm}
\includegraphics[width=\columnwidth]{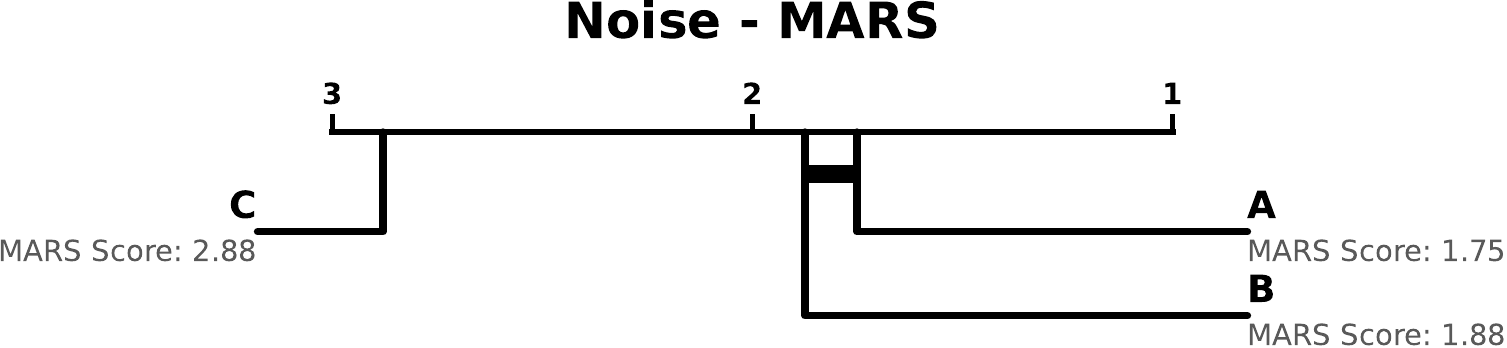}
\captionof{figure}{Comparison for Scenario 4: Standard (Top) vs.~MARS (Bottom).}
\label{fg:sc4}
\end{table}

\subsection{Scenario 4: Noisy Superiority}
This scenario evaluates the noisy superiority phenomenon where a model wins by a noise-level margin ($0.0001$) in most cases but suffers significant performance degradation elsewhere. The details of this scenario are shown in Table~\ref{tbl:sc4}, and the resulting CD diagrams are presented in Figure~\ref{fg:sc4}.

\textbf{Analysis:} Standard CD diagrams show Method B as the leader, which is arguably misleading. MARS reveals that Method B's rank is inflated by insignificant wins, showing Method A as the statistically superior choice once magnitude is considered.

\begin{table}[tb!]
\centering
\caption{Average Accuracy for Scenario 5}
\label{tbl:sc5}
\begin{tabular}{lcc}
\hline
\textbf{Classifier} & \textbf{$D_{0-29}$ (Easy Tasks)} & \textbf{$D_{30-39}$ (Edge Case)} \\ \hline
Method A            & 0.950                            & 0.940                            \\
Method B            & 0.951                            & 0.400                            \\
Method C            & 0.952                            & 0.350                            \\ \hline
\end{tabular}

\vspace{0.5cm}

\includegraphics[width=.82\columnwidth]{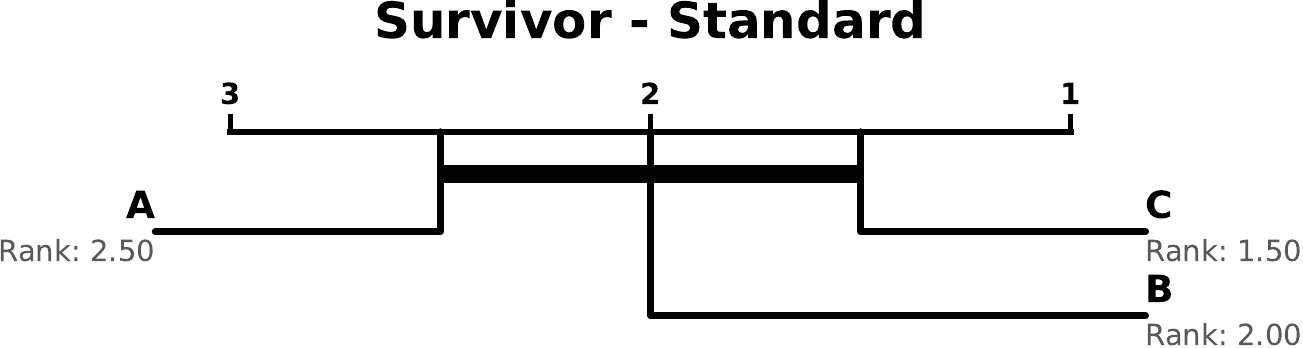} \\
\vspace{0.5cm}
\includegraphics[width=\columnwidth]{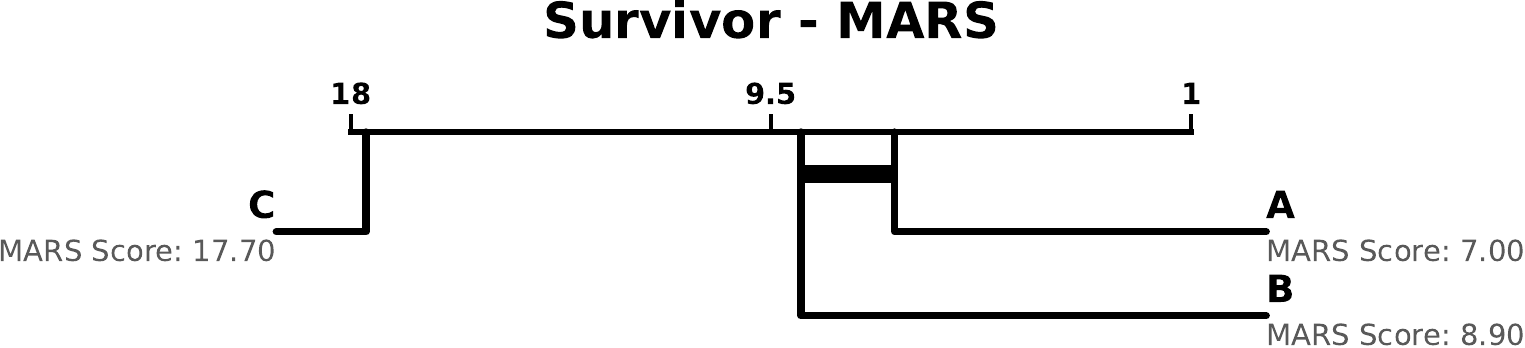}
\captionof{figure}{Comparison for Scenario 5: Standard (Top) vs.~MARS (Bottom).}
\label{fg:sc5}
\end{table}

\subsection{Scenario 5: The Survivor}
This scenario identifies the Survivor model. Most models perform well until a difficult edge case occurs, where only one model (A) maintains acceptable performance levels. The details of this scenario are shown in Table~\ref{tbl:sc5}, and the resulting CD diagrams are presented in Figure~\ref{fg:sc5}.

\textbf{Analysis:} Standard ranking prefers B and C due to their higher win count on easy data. MARS isolates Method A as the clear winner, as the weighting mechanism heavily penalizes the failure of B and C to remain competitive in edge-case scenarios.
\begin{table}[tb]
\centering
\caption{Base Accuracy for Scenario 6. Gaussian Noise ($\mu=0, \sigma=0.05$) per dataset, and Gaussian Noise ($\mu=0, \sigma=0.02$) per method.}
\label{tbl:sc6}
\begin{tabular}{lc}
\hline
\textbf{Classifier} & \textbf{Base Accuracy} \\ \hline
Method A    & 0.92                  \\
Method B            & 0.90                  \\
Method C            & 0.89                  \\
Method D            & 0.85                  \\
Method E            & 0.82                  \\
Method F            & 0.78                  \\
Method G            & 0.75                  \\
Baseline            & 0.60                  \\ \hline
\end{tabular}
\vspace{0.5cm}

\includegraphics[width=.88\columnwidth]{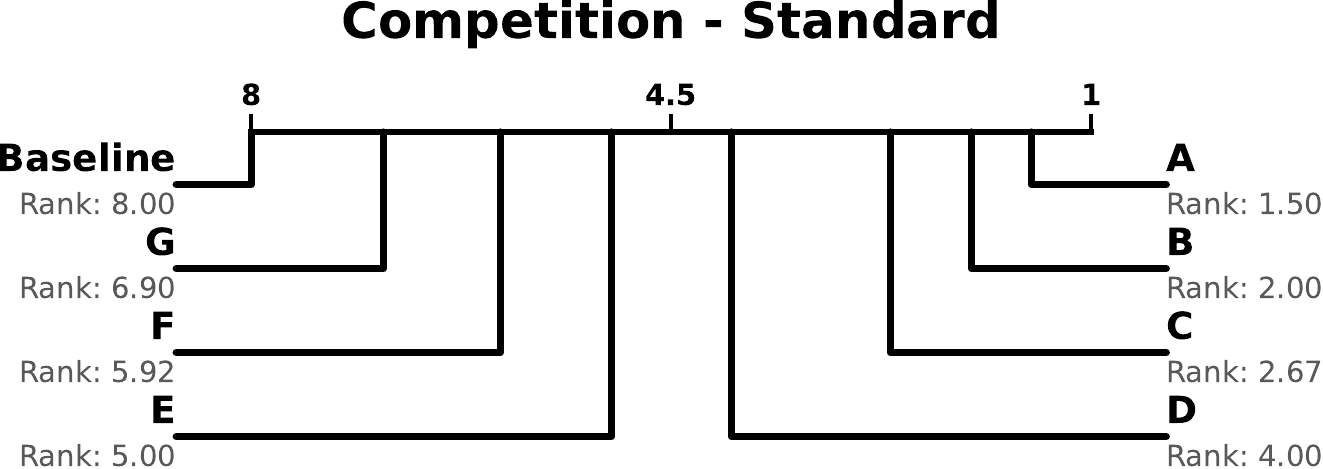} \\
\vspace{0.5cm}
\includegraphics[width=\columnwidth]{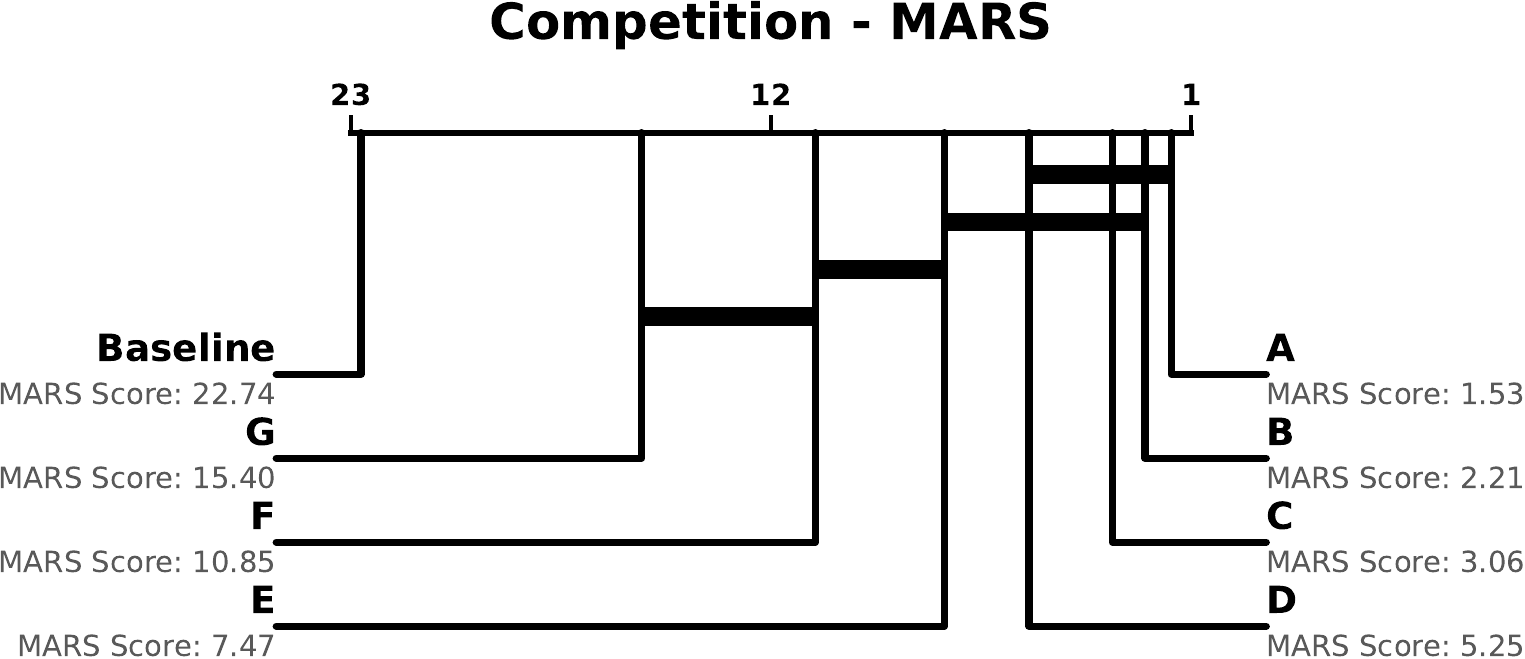}
\captionof{figure}{Comparison for Scenario 6: Standard (Top) vs. MARS  (Bottom).}
\label{fg:sc6}
\end{table}

\subsection{Scenario 6: Realistic Competition}
In this scenario, we aim to simulate a robust experimental setup across 40 datasets. To ensure a realistic scenario and to avoid static results, the base accuracies in Table~\ref{tbl:sc6} are subjected to two layers of stochastic variation:
\begin{enumerate}
    \item \textbf{Dataset Difficulty:} A global noise  ($\mu=0, \sigma=0.05$) based on the Gaussian distribution is added to the performance metric value for all models per dataset, simulating the inherent difficulty variance across different data distributions.
    \item \textbf{Model-Specific Noise:} A secondary Gaussian noise term ($\mu=0, \sigma=0.02$) is applied to each individual model's performance, allowing for occasional rank swaps that mirror the inconsistencies found in real-world cross-validation.
\end{enumerate}

This simulation provides a realistic competition scenario between 8 different methods on 40 datasets. We deliberately select one of the methods as a baseline, which is highly under-performing compared to the other methods, to have an obvious inferior candidate.

\textbf{Analysis:} The Standard rank statistics, as reflected in the CD diagram in Figure~\ref{fg:sc6}, produces the ranking as expected. It successfully identifies the underlying hierarchy of the base accuracy, but as it is not incorporating the performance metric values in the calculations, it only relies on discrete ranks and considers every method statistically different from each other. MARS on the other hand, offers a different perspective with more information reflection. The average ranks hierarchy provided by MARS is similar to the standard rank statistics, however, it's view on statistical significance is different.

With respect to MARS, Method A and D do not show a significant statistical difference. MARS uses metric values and the exact magnitude of difference to calculate rank scores and critical difference values. In the case of A and D, A has a base accuracy of $0.92$, and $D$ has a base accuracy of $0.85$. With respect to the $0.05$ dataset-wise Gaussian noise and $0.02$ method-specific Gaussian noise, in worst-case (or best-case) scenario, A might show a performance metric value of $0.85$, and D might yield a performance metric value of $0.92$, basically swapping their performance levels. Of course as the distribution is Gaussian, the probability of this event is low. MARS successfully reflects that in its analysis. It shows a quite high margin of difference between A and D (rank score of $1.53$ vs $5.25$ respectively), while it is also reflecting that there is a probability that these methods are actually very similar. We can also observe that the penalization becomes higher as we move towards the worst performance. For instance, E and G are in the same pairwise situation as A and D, but as G is very close to the worst performance, it is rightfully not included in the same critical difference. 

This phenomenon suggests that adding more datasets is likely to change the average ranking of the methods. This observation suggests that MARS is likely to be less sensitive to the number of observations, reflecting more information from the metric values compared to the standard rank analysis.

\subsection{Discussion}
We demonstrated six scenarios in which standard ranking statistics completely or partially breaks down and fails to provide valid and realistic insights on the evaluation results. Our suggested Magnitude-Aware Rank Statistics (MARS), on the other hand, successfully captures important properties of the experimental results and provides robust and reliable insights into the experimental results and comparisons. We also show that MARS can draw more information from limited observations, reflecting the landscape of the performance better, as it incorporates the performance metric values in the calculations.

It is noteworthy that MARS is still sensitive to the frequency of the rankings, and as more observations are included, analytical results based on MARS become more certain. As observed, MARS is inherently robust against extremely few anomalous observations in the performance metric values, and yields reliable and robust results for the ranking statistics.
Thus, as the unweighted standard ranking statistic, also MARS remains susceptible to the selection of datasets used for the performance test. Selecting meaningful datasets at a comparable and reasonable level of difficulty remains the responsibility of the researcher performing the test.

\section{Conclusion} \label{sec:conc}
In this paper, we proposed MARS, a magnitude-aware rank statistics and critical difference to facilitate more robust and realistic model evaluation. 
The traditional rank statistic has a strong advantage over a faulty procedure often followed earlier, namely simply averaging performance scores over many datasets, where the performance scores are actually not comparable. The ranks are indeed comparable, disregarding the scale of performance measures. However, the  magnitude-blindness of the traditional rank statistic can again lead to derivation of problematic and incorrect conclusions about the comprehensive evaluation results. We showed how MARS can help fixing the issues of traditional rank analysis in different edge cases and scenarios in taking the scale into account again, however, not directly but with a built-in normalization by using relative scores. 

An easy-to-use software package is also provided to easily conduct MARS, as well as the standard rank statistics and analysis, and to generate Critical Difference (CD) diagrams.

MARS in general is able to provide more information as it takes the performance metric values into account directly. We observed through different scenarios that the picture which MARS provides often includes more insights and leads to a deeper understanding of the entire evaluation and experimental results. The incorporation of performance metric values and avoiding \emph{magnitude-blindness} helps making the experiments and analyses more robust, often offering a more comprehensive view of the experimental landscape.

The final question is whether the standard rank analysis and statistics should be entirely replaced by MARS or not. We would like to state and emphasize, that many decisions with regards to the experiments and the tests are the responsibility of the researcher. Selecting suitable datasets, carefully evaluating the competitors, choosing the evaluation framework, and other decisions as such, are very important for conveying the right message which lays within the performance of a model and the experimental results. In this paper, we proposed MARS and highlighted the reasons which make us believe it is a proper and insightful way of conducting rank analyses. Using MARS instead of or alongside with the standard rank statistics, or not using it at all, is the decision and the responsibility of the researcher conducting the evaluations and tests.


\bibliography{myref}
\end{document}